\documentclass[conference]{IEEEtran}
\IEEEoverridecommandlockouts
\usepackage{cite}
\usepackage{amsmath,amssymb,amsfonts}
\usepackage{algorithmic}
\usepackage{multirow}
\usepackage{graphicx}
\usepackage{textcomp}
\usepackage{xcolor}
\usepackage{numprint}
\usepackage[bookmarks=false]{hyperref}
\def\BibTeX{{\rm B\kern-.05em{\sc i\kern-.025em b}\kern-.08em
    T\kern-.1667em\lower.7ex\hbox{E}\kern-.125emX}}
\begin{document}
\npdecimalsign{\ensuremath{.}}

\title{Deep Aggregation of Regional Convolutional Activations for Content Based Image Retrieval
}

\author{
Konstantin Schall, Kai Uwe Barthel, Nico Hezel, and Klaus Jung\\
\textit{Visual Computing Group, HTW Berlin}\\
Berlin, Germany\\ 
{\{schallk, barthel, hezel, jungk\}@htw-berlin.de}
}


\maketitle

\begin{abstract}
One of the key challenges of deep learning based image retrieval remains in aggregating convolutional activations into one highly representative feature vector. Ideally, this descriptor should encode semantic, spatial and low level information. Even though off-the-shelf pre-trained neural networks can already produce good representations in combination with aggregation methods, appropriate fine tuning for the task of image retrieval has shown to significantly boost retrieval performance. In this paper, we present a simple yet effective supervised aggregation method built on top of existing regional pooling approaches. In addition to the maximum activation of a given region, we calculate regional average activations of extracted feature maps. 
Subsequently, weights for each of the pooled feature vectors are learned to perform a weighted aggregation to a single feature vector. 
Furthermore, we apply our newly proposed NRA loss function for deep metric learning to fine tune the backbone neural network and to learn the aggregation weights. Our method achieves \mbox{state-of-the-art} results for the INRIA Holidays data set and competitive results for the Oxford Buildings and Paris data sets while reducing the training time significantly.
\end{abstract}


\section{Introduction}
With the rising amount of visual data produced and shared on the internet, the task of grouping and searching for related content becomes more challenging and more important. Information about millions of images has to be stored in very compact representations to be able to efficiently accomplish such tasks. Recent works have shown that representations extracted by a deep neural network can be used to form excellent feature vectors, which encode visual similarity of images in a high dimensional space \cite{Babenko14,Tolias16, Gordo17}. 
A convolutional neural network (CNN) often maps visually similar images to similar feature vectors whereas the extracted feature vectors are different for dissimilar images. Since these feature vectors are represented in floating point numbers, they can be easily compared by simple distance metrics such as $L^2$, which makes them suited for a wide range of instance level image retrieval tasks. Most neural networks used for image retrieval however, have been trained to classify images into a fixed number of categories. Even though these networks already show good retrieval results \cite{Seddati17}, performance can be improved by applying a model that has been specifically trained to rank images by their similarity as shown in \cite{Gordo17} and \cite{Radenovic17}.

Deep metric learning forms a sub-field of deep learning, where the goal is to find a representation space, in which visual resemblance of images is perfectly encoded by spatial relationships in the feature vector space. In these approaches, optimization occurs directly on the extracted embedding vectors. Distances between samples of the same class are required to be smaller than distances between entities of different classes \cite{Schultz03, Song16, Sohn16, Song17}. Metric learning methods therefore use loss functions, which are trying to teach a neural network to generate such a representation space. Networks trained in this way are better suited for the task of image retrieval \cite{Song16}. The Triplet loss function \cite{Schultz03} is the most prominent example. 
\IEEEpubidadjcol
In \cite{Gordo17} the authors achieve excellent retrieval results by building an end-to-end trainable system which extracts regional maximum activations from convolutional feature maps similar to R-MAC \cite{Tolias16}. They follow the ideas described in \cite{Tolias16} and aggregate these activations to one high dimensional feature vector. The network is trained with the Triplet loss function \cite{Schultz03} using the landmarks data set from \cite{Babenko14} in one of several training steps. The authors scale images to 800 pixels at the larger dimension during training, which makes the whole process very computationally expensive due to GPU memory limitations. 

In \cite{NRA19} we proposed the concept of \textit{Nonlinear Rank Approximation} (NRA) loss. This loss function achieves very good results and outperforms the Triplet loss on a wide range of retrieval tasks significantly. In this paper we propose to apply the NRA loss function to train an end-to-end retrieval network similar to \cite{Gordo17}. Since the NRA loss works best if batches are built from several classes containing multiple samples, large images cannot be used during training. However, we still can use relatively large image sizes for feature extraction due to the fully convolutional nature of the network and achieve better retrieval results with smaller resolutions than used in comparable work. Furthermore, we argue that the sum aggregation of the chosen number of regions extracted with the R-MAC method \cite{Tolias16} does not fully take the scale and size of different regions into account. Instead we apply a small aggregation network on top of the backbone CNN, which learns the importance of each of the regions and assigns it an individual weight. The total system is trainable in under 24 hours on a single NVIDIA GTX 1080 Ti GPU with 12 GB memory, where 64 images fit into one single batch using the ResNet50 V2 architecture \cite{He15} as backbone CNN.

\section{Related Work}
In this section we first describe the principle of the R-MAC descriptor, followed by an explanation of similar end-to-end trainable deep learning based image retrieval methods. Finally we explain the importance of the loss function, which is used to train the retrieval network. 

\subsection{The R-MAC descriptor}
In \cite{Tolias16} Tolias et al. describe a way to incorporate spatial information extracted with a convolutional neural network into an image's feature vector. Usually the last convolutional layer of the chosen backbone CNN is used to obtain a set of feature maps $\mathcal{X} = \{\mathcal{X}_i | i = 1...\mathcal{C} \}$, where $\mathcal{C}$ is the number of output channels in the said layer and the final feature vector will have $\mathcal{C}$ dimensions. These feature maps have been aggregated in various ways prior to the R-MAC method as in \cite{Girshick13, Azizpour14, Babenko15} neglecting the spatial information available. In \cite{Tolias16} a grid structure of variable scales is defined, consisting of 20 regions for three different scales $l = 1 \ldots 3$. The region's size for scale $l$ is calculated by $2\min(W, H) / (l + 1)$. The regions are uniformly chosen in a sliding window manner with the current scale. In the next step, the maximum activation value of each region and feature map is chosen to form 20 $\mathcal{C}$-dimensional feature vectors. These are furthermore $L^2$-normalized, whitened with a PCA \cite{Jegou12} and then $L^2$-normalized again individually, before they are sum pooled to produce a single vector, the R-MAC descriptor.

\subsection{End-to-End Networks} 
In \cite{Gordo17} the Deep Image Retrieval (DIR) system is described which is based on the previously mentioned R-MAC approach \cite{Tolias16}. The authors goal was to design a convolutional feature extractor pipeline as presented in \cite{Tolias16} and train the resulting architecture in an end-to-end manner, since all applied operations are differentiable. Importantly, the authors claim that the Softmax Cross Entropy loss function is not optimal for the task of fine tuning for retrieval and apply the Triplet loss function \cite{Schultz03} instead. However, since the used landmarks data set \cite{Babenko14} seems to be noisy, i.e. images of the same category are visually different, they could not train the proposed network with the Triplet loss directly. Instead they first fine tune the used ResNet101 \cite{He15} in a traditional, classification approach and then perform a cleaning of the landmarks data set, based on a visual matching comparison of SIFT image descriptors \cite{Lowe03}. This cleaning reduces the landmarks data set from \numprint{192000} to \numprint{42410} training images with 586 categories. The authors refer to this version of the data set as \textit{Landmarks Clean}.
Furthermore, as retrieval performance seems to increase for input images with larger sizes \cite{Gordo17, Magliani18}, Gordo et al. decide to perform the last step of their training procedure, i.e. fine tuning with Triplet loss with the landmarks clean data set, with images of relatively large size. Specifically, each training image is resized to 800 pixels at the larger side. However, one single image of this size results in approximately 7.5 GB GPU memory, which limits training to one image per batch on most GPUs. To overcome this limitation the authors introduce a modified sequential training procedure. With this modification, they are indeed able to train the proposed architecture and achieve outstanding retrieval results at the price of one week of total training time. At the end of the training, the authors further improve their models performance by extracting feature vectors from images in multiple resolutions, namely 550, 800 and 1050 pixels at the larger side. Those feature vectors are $L^2$-normalized, sum aggregated and $L^2$-normalized again.

Noh et al. present DELF (DEep Local Feature) in \cite{Noh17}. A pre-trained ResNet50 network is used as the backbone CNN and two additional convolutional layers are added to model an attention functionality. The specific task of this part is to analyze the extracted set of feature maps and to assign an importance weight to each area. Images of size $224 \times 224 \times 3$ are used, which leads to feature maps of shape $7 \times 7 \times 2048$. Each of the 49 2048-dimensional feature vectors corresponds to a part in the input image which is specified by the network's receptive field \cite{Le17}. The attention part of the DELF architecture analyzes each of the 49 vectors and learns to assign a higher weight to a single area, if an interesting object is seen. For training, the authors further sum-aggregate the weighted combination of the 49 embeddings to form a single 2048-dimensional feature vector. Training is done in two steps. First, the original pre-trained ResNet50 is fine tuned with the same landmarks set as in \cite{Gordo17}. In the second step the attention part is trained independently, i.e the weights of the backbone CNN are not further optimized. The Softmax Cross Entropy loss function is used during both training procedures.

\subsection{Loss Functions for Retrieval}
\begin{figure}
    \centering
    \newlength\imgheight
    \setlength\imgheight{0.22\textwidth}
    \begin{tabular}{cc}
    \includegraphics[height=\imgheight]{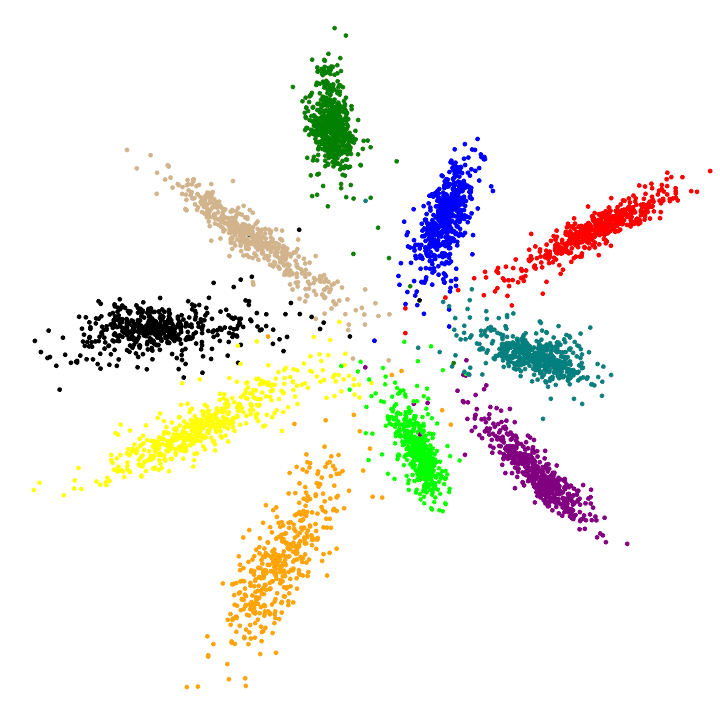} & 
    \includegraphics[height=\imgheight]{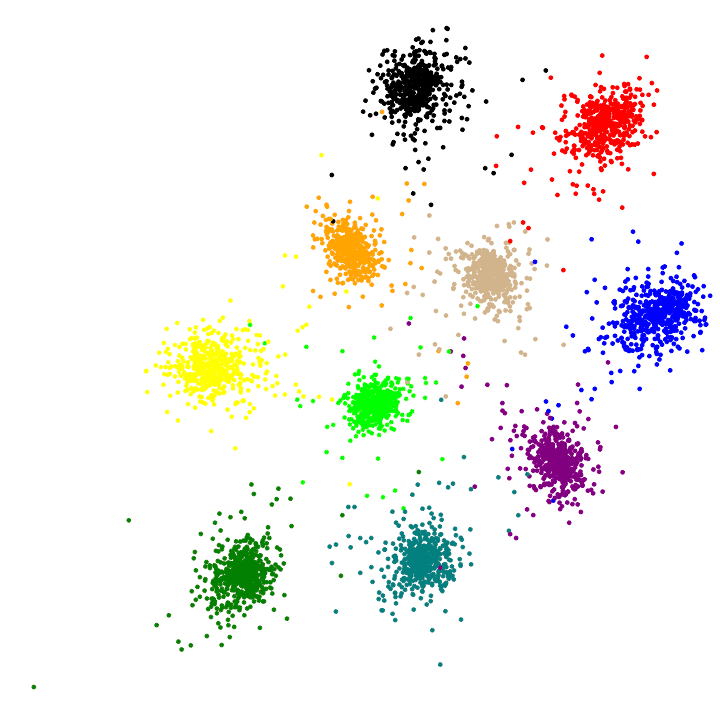} \\
    (a) Softmax Cross Entropy & \quad (b) Triplet
    \end{tabular}
    \caption{Visualization of the 2D embedding space of MNIST numbers \cite{LeCun98} after training the same network with two different loss functions. Representations produced by the Softmax Cross Entropy loss are not as suited for nearest neighbor search as the ones produced by the Triplet loss.}
    \label{fig:embedding}
\end{figure}

For an image classification task the set of feature maps $\mathcal{X}$ is reduced to a $\mathcal{C}$-dimensional vector by averaging each activation map $\mathcal{X}_i$. This vector is further fed through a fully connected and a Softmax layer, where both have outputs equal to the number of classes of the training set. The output of the Softmax function can be seen as a probability distribution of the available classes. The Cross Entropy loss then tries to minimize the divergence between the ground truth and the predicted Softmax output. To achieve a low Softmax Cross Entropy, it is sufficient to divide the average pooled CNN representation space into as many regions as there are classes in the training (see Fig. \ref{fig:embedding} (a)). 
However, such a constellation is not suited for nearest neighbor search. 

Since image retrieval quality is evaluated by the ranking order of nearest neighbors to a query image, the representation space should form tighter clusters for optimal search results. One loss function trying to achieve this goal is the Triplet loss function \cite{Schultz03}, which builds a set of three samples $\mathcal{T} = \{\boldsymbol{x}^a, \boldsymbol{x}^+, \boldsymbol{x}^-\}$  consisting of an anchor  $\boldsymbol{x}^a$, one positive example $\boldsymbol{x}^+$ of the same class as the anchor and one negative example  $\boldsymbol{x}^-$ of a different class. 
 
The loss value for a single triplet is calculated by Eq. (\ref{eq:TripletLoss}) where $D^2$ denotes the $L^2$-distance between two feature vectors $f(\boldsymbol{x})$ with
$D_{i,j} = \| f(\boldsymbol{x}_i) - f(\boldsymbol{x}_j) \|_2$.
\begin{equation}
\label{eq:TripletLoss}
J_\mathcal{T} = \max{(0,D_{a,+}^2 - D_{a,-}^2 + \alpha})
\end{equation}

The Triplet loss tries to maximize the relative distance to a negative example and minimizes the distance to positive ones, ideally grouping images from the same class to clusters in the given representation space.

\section{Proposed Method}
\subsection{Nonlinear Rank Approximation Loss}
Following the assumption that a perfect clustering is achieved if and only if all distances to negative examples are larger than the maximum distance to positive examples, we introduced the Nonlinear Rank Approximation loss (NRA) in \cite{NRA19}. The main principles of this loss function will be briefly explained below. 

NRA is taking the two most important distances w.r.t. an anchor into account: The maximum distance to the embedding of its positive examples of the same class and the minimum distance to its negative examples from a different class:
\begin{equation}
D^+_{i,\max} = \max_{j} D_{i,j} \quad D^-_{i,\min} = \min_{j} D_{i,j}
\end{equation}

To enable a search for those points, $m = k \cdot n$ samples $\boldsymbol{x}_i$ of the input data are used per batch, where $k$ is the number of classes and $n$ the number of samples per class. This step ensures each sample in the batch to have $n-1$ positive and \mbox{$m-n-1$} negative examples.
For each feature vector in the batch, the distances to all other feature vectors are determined. This yields a $m \times m$ distance matrix, where $D_{i,\max}$ is the maximum of its $i^{\text{th}}$ row values and $D_{i,\min}$ is the minimum accordingly. 
\begin{equation}
 D_{i,\max} = \max_{j} D_{i,j} \quad D_{i,\min} = \min_{j}  D_{i,j}
\end{equation} 

Instead of using this distance matrix directly, the approximated ranks $r_{i,j}$ are computed by min-max-scaling each row to the interval $[0,1]$ individually: 
\begin{equation}
r_{i,j} = \frac{D_{i,j} - D_{i,\min}}{D_{i,\max} - D_{i,\min}} \in [0, 1]
\label{eq:normalized-rank}
\end{equation}

This scaling ensures a unified representation of the distances for each batch during training by approximating a ranking order, where the lowest possible rank is 0 and the highest is 1. 
Subsequently, the approximated normalized ranks for the most distant embedding of the same class $r^+_{i,\max}$ and the closest embedding of a sample of a different class $r^-_{i,\min}$ are determined by
\begin{equation}
r^+_{i,\max} = \frac{D^+_{i,\max} - D_{i,\min}}{D_{i,\max} - D_{i,\min}} \quad r^-_{i,\min} = \frac{D^-_{i,\min} - D_{i,\min}}{D_{i,\max} - D_{i,\min}}
\end{equation}

Next for each row these two ranks are transformed to similarities. Instead of calculating the similarity as $s = 1 - r$  a parametric nonlinear transfer function is used
\begin{align}
w(\cdot;\alpha)&: [0, 1] \rightarrow [0, 1] \\
w(r; \alpha) &= 
\begin{cases}
\frac{1}{2}(2r)^\alpha & r \in [0, \frac{1}{2}) \\
1 - \frac{1}{2}(2(1-r))^\alpha & r \in [\frac{1}{2}, 1] \\
\end{cases}
\label{eq:nonlinearity}
\end{align}

with $\alpha\in\mathbb{R}^+$. The parameter $\alpha$ controls the slope of $w(\cdot;\alpha)$ at $r = 0.5$. Thus, it determines the influence of rank errors. With increasing $\alpha$ falsely ranked negative images with a high similarity are stronger punished and correctly ranked positive images with a high similarity will result in a loss value closer to zero. An empirical study showed best results for $\alpha=4$ \cite{NRA19}.  
Using $w(\cdot;\alpha)$ and omitting $\alpha$ in the following notation we define the similarity of a sample $\boldsymbol{x_j}$ to the anchor $\boldsymbol{x_i}$ by $s_{i,j} = 1 - w(r_{i,j})$. The final loss value for each batch is calculated as follows:  
\begin{equation}
J = - \frac{1}{m} \sum_{i = 1}^m \left( \log(s^+_{i,\max} + \varepsilon) + \log(1 - s^-_{i,\min} + \varepsilon) \right)
\end{equation}

Here $\varepsilon > 0$ is a small number that controls the maximum loss contribution if $s^+_{i,\max}=0$ or $s^-_{i,\min}=1$. We use $\varepsilon = 10^{-4}$ throughout the following experiments. 

\subsection{Global and Regional Pooling}

The R-MAC approach \cite{Tolias16} aggregates maximum activations of different regions in convolutional feature maps into one final feature vector per image. However, the maximum activation of a region can be very high, even though the detected shape is only seen on a small area of the region. Furthermore, since many of the 20 regions have overlapping areas, the pooled max value will be the same in some of the areas. Calculating the average activation value of each region gives a more accurate representation of an image's convolutional activations but is more sensitive to changes in the input image's size. To better understand the differences of both pooling operations, we conduct a preliminary study and build different feature vectors from convolutional activation maps extracted with a ResNet50 v2 network \cite{He16}  for different input image sizes. The network is pre-trained with images with 224 pixels on both sides from the ILSVRC2012 data set \cite{Russakovsky14} and we use images from a rotated version of the INRIA Holidays data set \cite{Holidays} for evaluation. 

\begin{figure}
\centerline{\includegraphics[width=0.5\textwidth]{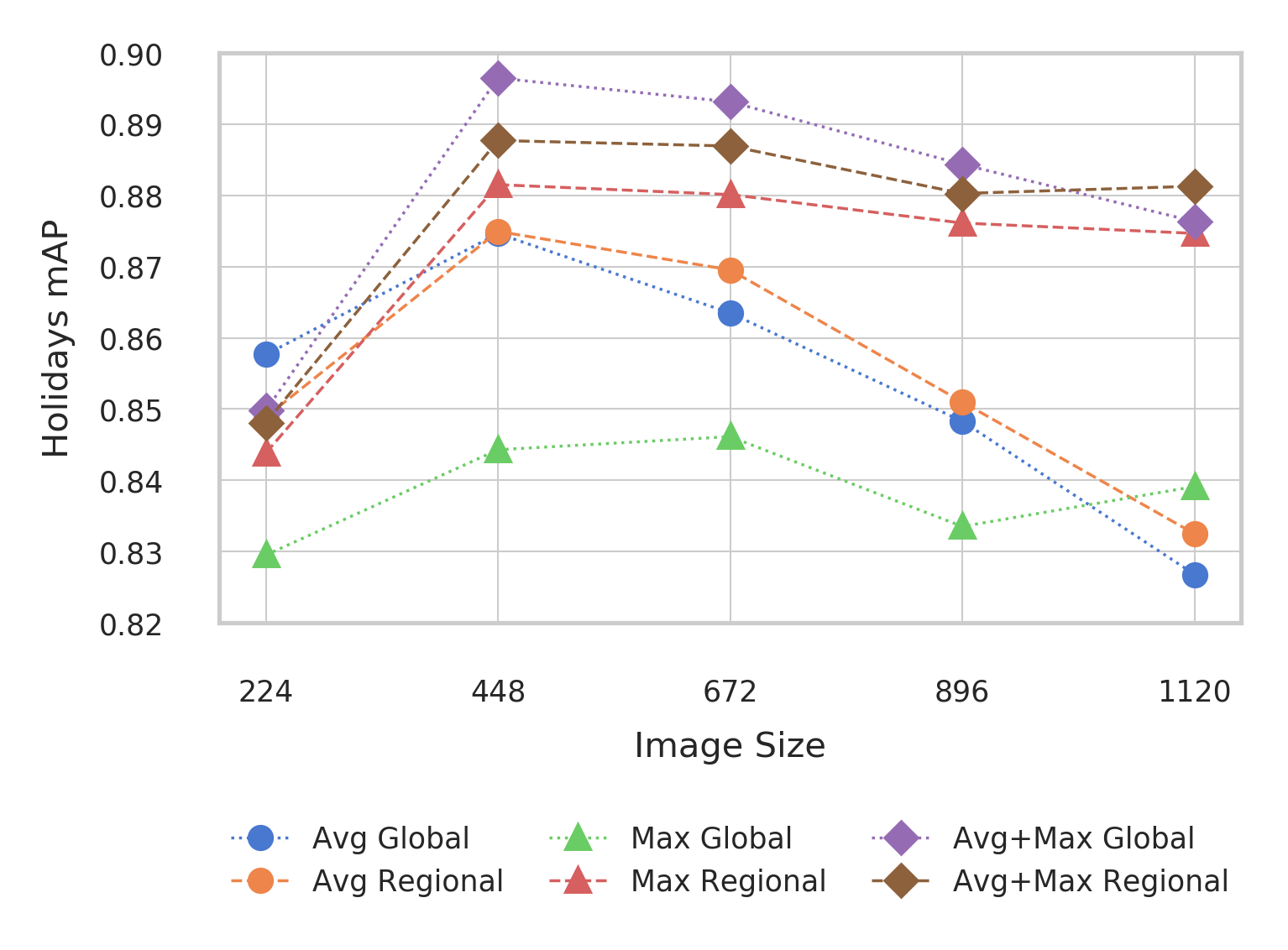}}
\caption{Mean Average Precision scores for the Holidays data set and different aggregation techniques for varying input image sizes.}
\label{fig:Prestudy}
\end{figure}

The networks last convolutional layers are used as a feature map extractor and six different pooling techniques are compared in the preliminary study. The first is built by a global average pooling of the entire set of feature maps (Avg Global), whereas the second comes from the global max pooling operation (Max Global). The third represents the sum of those two globally pooled values (Avg+Max Global). Further, we build three regional pooled and aggregated feature vectors. One for the average pooled regional activations \hbox{(Avg Regional)}, another for the max pooled regional activations (Max Regional) similar to R-MAC \cite{Tolias16} and the last is created by a pooling of maximum and average values from each region (Avg+Max Regional). For the regional approaches we choose the 20 areas as used in \cite{Tolias16}. In case of the combined regional approach, we obtain 20 max pooled and 20 average pooled values. Furthermore, we $L^2$ normalize each feature vector individually in the regional methods, sum pool over the regions and $L^2$ normalize again. With each of those six methods, we extract feature vectors for five image sizes $S \in \{224, 448, 672, 896, 1120\}$. Fig. \ref{fig:Prestudy} summarizes the results from this experiment. 

Several important conclusions can be drawn from these results. Firstly, all aggregation methods benefit from an image size, twice as high as the images seen during training. Global max pooling performs poorly across all tests. The mean average precision scores drop drastically if using average pooling based methods for higher resolutions. The R-MAC method achieves good results, but is outperformed by the pooling techniques that use a combination of max and average pooling. 

\begin{figure}
\centerline{\includegraphics[width=0.5\textwidth]{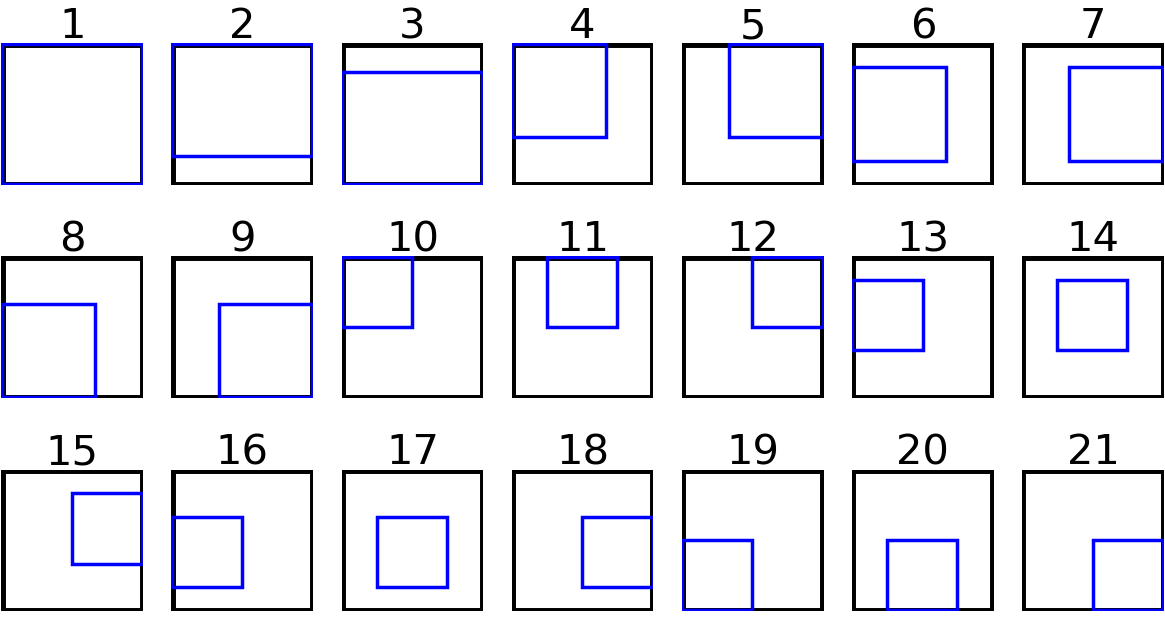}}
\caption{The 21 regions used for regional max and average pooling}
\label{fig:grid}
\end{figure}

In this experiment, the global version of the combined max and average pooling method (Avg+Max Global) outperforms the regional approach (Avg+Max Regional) in most resolutions. We argue that this is due to the fact that the 40 pooled values are not equally important, but are handled this way during the sum aggregation. For example, the activations of individual regions should be assigned a weight according to their size and the average value seems to be much more important in the bigger regions compared to the maximum value. Furthermore, the global activation map should be added as a region, since the global max and average pooled feature vector yields such good results. This leads to a total of 21 areas (Fig. \ref{fig:grid}) and 42 pooled values, which will be aggregated in the following steps. 

\subsection{Weighted Aggregation of CNN Activations}
\begin{figure}
\centerline{
\includegraphics[width=0.5\textwidth,trim={0 7.5cm 10cm 0},clip]{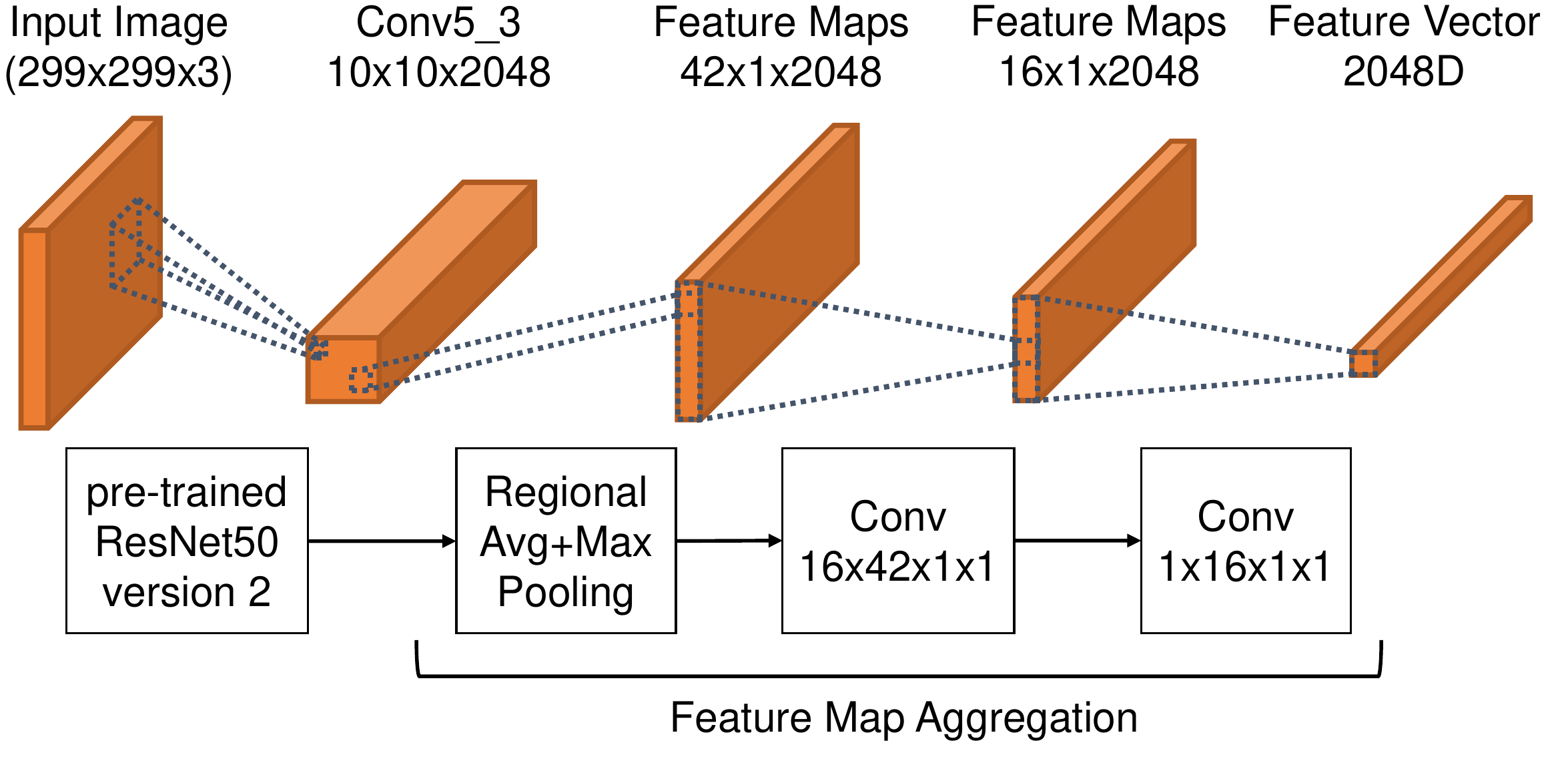}
}
\caption{Functionality of the proposed DARAC System. A two layer convolutional layer with a very small number of trainable parameters is used to aggregate maximum and average activations from different regions of feature maps produced by a CNN.}
\label{DARAC}
\end{figure}

Performing regional max and average pooling with a grid structure as seen in Fig. \ref{fig:grid} on sets of feature maps produced by the last convolutional layer with a ReLU activation, results in a matrix of dimension $42 \times C$ with values in $\mathbb{R}_{\ge 0}$, where $C$ is the number of kernels in the last layer of the chosen CNN. Following the conclusions from the previous experiment, we design a convolutional approach to assign a weight to each of these values (Fig. \ref{DARAC}). We refer to this method as DARAC (Deeply Aggregated Regional Activations of Convolutions). In the first step a convolution layer with $l$ kernels of size $42 \times 1$ aggregates the pooled data to a matrix of dimension $l \times C$. The output of this operation is followed by a ReLU activation function, which pushes most learned kernel coefficients to be at least 0. The $l$ row vectors of the obtained matrix are further batch normalized \cite{Szegedy15} to get mean centered feature vectors with unit variance, a representation similar to a PCA whitening output \cite{Jegou12} as used in \cite{Tolias16} and \cite{Gordo17}. Those values are further convoluted with a single $l \times 1$ kernel to output one $C$-dimensional feature vector. This vector representation is then used as the embedding space to optimize with the described nonlinear rank approximation loss function without any normalization or post processing during the training procedure. The two aggregation layers result in a very small overhead and only add $43l + l + 1$ (incl. bias) learnable parameters to the CNN architecture. We choose $l=16$ in the following experiments.

After convergence we use a post processing pipeline similar to \cite{Tolias16}. Firstly, the extracted feature vector is $L^2$ normalized and then we perform PCA whitening \cite{Jegou12}, where the principal components are determined on a subset of the training data set. The whitened vector is subsequently $L^2$ normalized again.  

\section{Experiments}
\subsection{Training}
We chose the ResNet50 v2 architecture \cite{He16} as the backbone CNN for the following experiments. Network weights are initialized with pre-trained parameters achieving a \mbox{top-1} accuracy of \numprint{76.47}\% on the ImageNet validation data set \cite{Russakovsky14}. The proposed aggregation part is connected to the network's last convolutional layer with 2048 kernels. 
The entire system is trained on the \textit{Google Landmarks} data set \cite{Noh17}. We choose $k=16$ classes and $n=4$ representatives for batch building. Therefore we build a subset of the landmarks data set, where all images from classes with less than four examples are excluded. This modification leads to a total of \numprint{1196934} images from \numprint{12602} classes. \numprint{100000} randomly chosen batches are processed in total and we use stochastic gradient descent with momentum as the optimizer with a learning rate of $\alpha = 0.0001$ throughout the entire training process. First all images are resized to 320 pixels on the smaller side and then they are randomly cropped to form a square with 299 pixels. These crops are subsequently randomly flipped horizontally for data augmentation purposes. Since the training on a cleaned landmarks data set appeared to be crucial for optimal retrieval results in \cite{Gordo17} we perform an additional training on a subset of the used collection with identical hyper-parameters after convergence. However, only \numprint{28326} images from 582 classes could be retrieved due to broken links. The resulting feature vectors from the first training procedure are further denoted as DARAC-GL and the vectors produced after the second training as DARAC-CL. We use the \textit{TensorFlow} framework \cite{Abadi16} for the DARAC implementation and all experiments are conducted on a single GeForce 1080 Ti GPU with 12 GB of memory.

\subsection{Evaluation}
Both network states are evaluated on three widely used retrieval data sets and we strictly follow the respective protocols for calculations of the mean average precision scores. The Paris \cite{Philbin08} and Oxford Building \cite{Philbin07} data sets consist of various landmarks, which makes them similar to the ones used during training. The Holidays data set \cite{Holidays} is more diverse and contains a mix of images other than landmarks. Similar to \cite{Radenovic17} and \cite{Gordo17}, we aggregate feature vectors extracted from different image scales. More specifically the images are resized to the resolutions of $S \in \{299, 540, 1020\}$. Each vector is post processed in the previously described manner individually and the whitened descriptors are sum aggregated and $L^2$ normalized subsequently. This approach is denoted as (\mbox{multi-resolution}) MR-DARAC. 

\subsection{Results}
Table \ref{table:DARAC} summarizes the retrieval results for the evaluation data sets for each of the described steps. As a baseline we show mAP scores for the global average pooled feature vector obtained from the pre-trained and trained ResNet50 network. The fine tuning on the Google Landmarks data set increases performance on all data sets but especially on Paris and Oxford Buildings, since those are very close to the training set. However, increasing the image size from 299 to 540 harms the results obtained with this average pooled layer, which corresponds with the previous findings. The introduced aggregation technique seems to be beneficial on its own but improves results even more, when increasing the image size and after a whitening of the feature vectors. Especially the multi-resolution approach seems to work very well on vectors produced by the introduced deep aggregation of regional convolutional activations. The second training procedure on the Landmarks Clean data set leads to improved results for the two buildings data sets but decreases the mean average precision for Holidays. Even though the authors of \cite{Gordo17} claim to exclude images of categories that occur in Paris and Oxford, the overall data set and category distribution appears to be closer to these data sets and therefore leads to better results.
\begin{table}
\centering
\caption{mAP scores for different stages of the proposed system}
\resizebox{\linewidth}{!}{
\begin{tabular}{c|c|c|c|c} 
 \hline
 Method & Image Size & Oxford & Paris & Holidays\\ 
 \hline\hline
 ResNet50 (pre-trained) & 299 & 48.7 & 68.9 & 86.1\\
 ResNet50-GL (trained) & 299 & 75.3 & 87.6 & 91.6\\
 ResNet50-GL (trained) & 540 & 72.5 & 85.3 & 90.9\\
 DARAC-GL & 299 & 77.6 & 89.4 & 92.8\\
 DARAC-GL + PCA{$_W$} & 299 & 81.4 & 90.8 & 93.7\\
 DARAC-GL + PCA{$_W$} & 540 & 82.2 & 91.9 & 95.5\\
 DARAC-GL + PCA{$_W$} & 1020 & 78.3 & 87.6 & 94.8\\
 MR-DARAC-GL + PCA{$_W$} & 299, 540, 1020 & 83.4 & 93.0 & \textbf{96.9}\\
 MR-DARAC-CL + PCA{$_W$} & 299, 540, 1020 & \textbf{88.2} & \textbf{94.1} & 95.5\\
 \hline
\end{tabular}
}
\label{table:DARAC}
\end{table}

\begin{table*}[ht!]
\centering
\caption{Comparison to similar content based image retrieval systems}
\begin{tabular}{ c|c|c|c|c|c|c } 
 \hline
 Method  & Matching & Network & Resolutions / Scales &  Oxford & Paris & Holidays\\ 
 \hline\hline
 DELF \cite{Noh17} & local & ResNet50 & 7, 0.25 - 2 & 83.8 & 85.0 & -\\
 GeM \cite{Radenovic17} & global & ResNet101 & 5, 0.25 - 1 & 87.8 & 92.7 & 93.9 \\
 DIR \cite{Gordo17} & global & ResNet101 & 550, 800, 1050 & 86.1 & \textbf{94.5} & 94.8\\  
 MR-DARAC-GL (ours) & global & ResNet50 &  299, 540, 1020 & 83.4 & 93.0 & \textbf{96.9}\\
 MR-DARAC-CL (ours) & global & ResNet50 & 299, 540, 1020 & \textbf{88.2} & 94.1 & 95.5\\
 \hline
\end{tabular}
\label{table:Sota}
\end{table*}

\subsection{Comparison with the state-of-the-art}
Table \ref{table:Sota} puts our method in comparison to similar deep learning based image retrieval systems. Training the proposed network architecture on Google Landmarks images leads to state of the art results for the rotated version of the Holidays data set. An additional fine tuning with the Landmarks Clean data set from \cite{Gordo17} results in descriptors that outperform the other methods for Holidays and Oxford. Only DIR \cite{Gordo17} achieves better results for the Paris data set. However, it should be noted that DIR and GeM \cite{Radenovic17} use a ResNet101. Furthermore, DIR spends a total of one week in training, whereas our system converges in under 24 hours on similar GPU hardware.

\section{Conclusion}
The presented DARAC system shows a very efficient method for aggregation of CNN features when fine tuned with the NRA loss for large scale image retrieval. Due to its fully convolutional nature it allows training with relatively small images but achieves great results when using larger resolutions during testing. Additionally, it shows the importance of regional average pooling, which has been ignored in previous works regarding deep learning based image retrieval. A weighted aggregation of average and max pooled regional activations in form of convolutional layers leads to state-of-the-art results while reducing the training and feature extraction time significantly, since images of smaller sizes can be used in both stages. At the same time our method introduces a negligible number of trainable parameters compared to the number of weights in the backbone CNN, while also keeping a very low memory footprint. 
\bibliographystyle{abbrv}
\bibliography{WRAMAC_Paper}
\end{document}